\documentclass[sigconf]{acmart}
\usepackage{amsmath,amsfonts}
\usepackage{algorithmic}
\usepackage{algorithm}
\usepackage{array}
\usepackage[caption=false,font=normalsize,labelfont=sf,textfont=sf]{subfig}
\usepackage{textcomp}
\usepackage{stfloats}
\usepackage{url}
\usepackage{verbatim}
\usepackage{graphicx}
\usepackage{colortbl}
\usepackage{color,xcolor}
\usepackage{booktabs}
\usepackage{multirow}
\usepackage{paralist}
\newcolumntype{H}{>{\setbox0=\hbox\bgroup}c<{\egroup}@{}}
\definecolor{red}{RGB}{202,0,0}
\definecolor{blue}{RGB}{0,102,204}
\definecolor{green}{RGB}{50, 205, 50}
\AtBeginDocument{%
  }

\copyrightyear{2025}
\acmYear{2025}
\setcopyright{acmlicensed}\acmConference[MM '25]{Proceedings of the 33rd
ACM International Conference on Multimedia}{October 27--31, 2025}{Dublin,
Ireland}
\acmBooktitle{Proceedings of the 33rd ACM International Conference on
Multimedia (MM '25), October 27--31, 2025, Dublin, Ireland}
\acmDOI{10.1145/3746027.3755355}
\acmISBN{979-8-4007-2035-2/2025/10}

\begin{document}

\title{ST-SAM: SAM-Driven Self-Training Framework for Semi-Supervised Camouflaged Object Detection}

\author{Xihang Hu}
\affiliation{%
  \institution{Jilin University}
  \city{Changchun}
  \country{China}}
\email{huxh24@mails.jlu.edu.cn}

\author{Fuming Sun}
\affiliation{%
  \institution{Dalian Minzu University}
  \city{Dalian}
  \country{China}}
\email{sunfuming@dlnu.edu.cn}

\author{Jiazhe Liu}
\affiliation{%
  \institution{Jilin University}
  \city{Changchun}
  \country{China}}
\email{liujz24@mails.jlu.edu.cn}

\author{Feilong Xu}
\affiliation{%
  \institution{Jilin University}
  \city{Changchun}
  \country{China}}
\email{xufl23@mails.jlu.edu.cn}

\author{Xiaoli Zhang}
\authornote{Corresponding author.}
\affiliation{%
  \institution{Jilin University}
  \city{Changchun}
  \country{China}}
\email{zhangxiaoli@jlu.edu.cn}

\renewcommand{\shortauthors}{Hu et al.}

\begin{abstract}
  Semi-supervised Camouflaged Object Detection (SSCOD) aims to reduce reliance on costly pixel-level annotations by leveraging limited annotated data and abundant unlabeled data. However, existing SSCOD methods based on Teacher-Student frameworks suffer from severe prediction bias and error propagation under scarce supervision, while their multi-network architectures incur high computational overhead and limited scalability. To overcome these limitations, we propose ST-SAM, a highly annotation-efficient yet concise framework that breaks away from conventional SSCOD constraints. Specifically, ST-SAM employs Self-Training strategy that dynamically filters and expands high-confidence pseudo-labels to enhance a single-model architecture, thereby fundamentally circumventing inter-model prediction bias. Furthermore, by transforming pseudo-labels into hybrid prompts containing domain-specific knowledge, ST-SAM effectively harnesses the Segment Anything Model's potential for specialized tasks to mitigate error accumulation in self-training. Experiments on COD benchmark datasets demonstrate that ST-SAM achieves state-of-the-art performance with only 1\% labeled data, outperforming existing SSCOD methods and even matching fully supervised methods. Remarkably, ST-SAM requires training only a single network, without relying on specific models or loss functions. This work establishes a new paradigm for annotation-efficient SSCOD. Codes will be available at \url{https://github.com/hu-xh/ST-SAM}.
\end{abstract}



\maketitle

\section{Introduction}
Camouflaged Object Detection (COD) aims to locate and segment camouflaged objects in complex scenes. The large size variations, complex contours, and similar textures of camouflaged objects render the COD task highly challenging and increase the need for pixel-level annotations. Nevertheless, the high cost of obtaining pixel-level annotated datasets significantly constrains the development and practical application of COD-related research.

\begin{figure}[t]
	\begin{center}
		\includegraphics[width=0.8\linewidth]{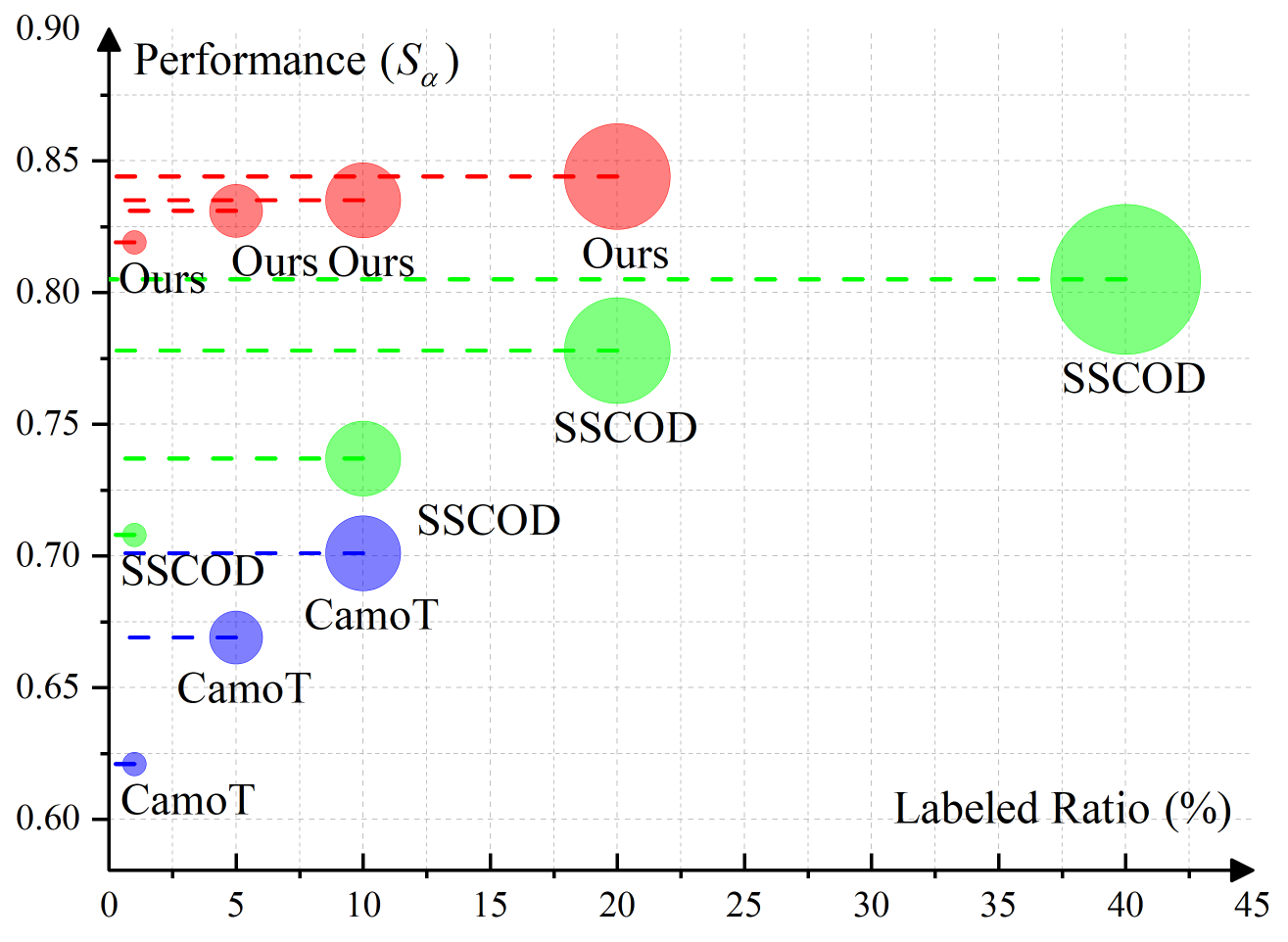}
	\end{center}
	\caption{Performance comparison of semi-supervised COD methods on the CAMO dataset under different proportions of labeled samples.}
	\label{fig:1}
\end{figure}

To mitigate the strong reliance on large-scale annotated data, Semi-Supervised Learning (SSL) \cite{ref45, ref46, ref47, ref48, ref49} has demonstrated outstanding performance in various fields. For instance, in Crowd Counting \cite{ref29} and medical image segmentation \cite{ref30}, SSL has achieved remarkable success by leveraging limited annotated data and abundant unannotated data. However, integrating SSL with COD presents unique challenges due to the intrinsic feature-space entanglement between highly camouflaged foreground objects and their backgrounds. Current SSCOD approaches \cite{ref6, ref7} adopt Teacher-Student frameworks with multi-network co-training. These methods face two critical limitations when labeled data is scarce: (1) the complex multi-network architecture tends to overfit, resulting in poor generalization to unlabeled samples; and (2) increasing prediction discrepancies between networks lead to severe error propagation. These factors collectively create strong dependencies on annotation quantity, making it difficult to balance performance and annotation efficiency. Furthermore, the multi-network paradigm requires substantial design and training overhead, and its inherent inter-network compatibility requirements significantly constrain scalability. To overcome these limitations, we propose to transcend the conventional Teacher-Student framework and develop a more concise strategy that reduces reliance on annotated data.

\begin{figure}[t]
	\begin{center}
		\includegraphics[width=0.9\linewidth]{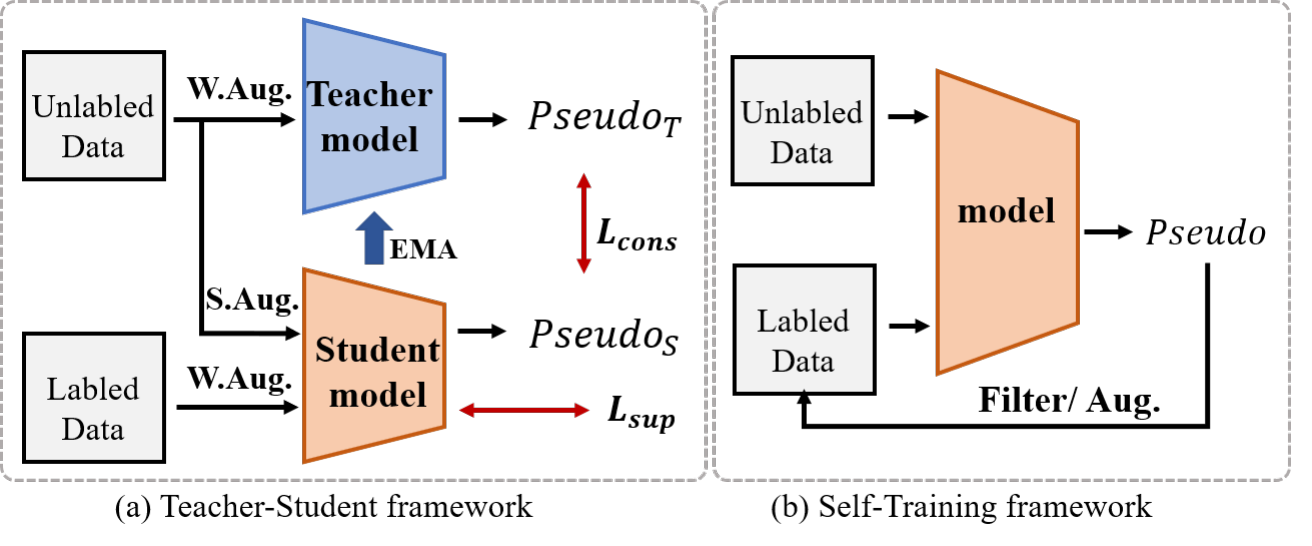}
	\end{center}
	\caption{The comparison of typical SSCOD frameworks: (a) Teacher-Student framework; (b) Self-Training framework. }
	\label{fig:a}
\end{figure}

Unlike the teacher-student frameworks, Self-Training strategies \cite{ref15, ref31, ref33} fundamentally expand the training set through pseudo-labels generated from unlabeled data to progressively refine the model (As shown in Fig. \ref{fig:a}). This approach circumvents the severe prediction bias inherent in multi-network architectures when labeled data is scarce by reinforcing a single network. Moreover, its model-agnostic nature grants exceptional scalability, as it imposes no constraints on specific architectures or loss designs. However, conventional self-training often struggles with complex tasks. The inherent complexity of COD leads to noise-contaminated pseudo-labels during initial phases, where error accumulation becomes progressively amplified during subsequent training, ultimately causing performance collapse. To harness the advantages of self-training, we aim to effectively utilize these noisy pseudo-labels by enabling their self-correction, thereby enhancing the strategy's adaptability to complex tasks.

Recent advances in visual foundation models, particularly the Segment Anything Model (SAM) \cite{ref5}, have demonstrated remarkable segmentation capabilities across diverse scenarios, presenting a promising solution. Nevertheless, in specific domains like COD, the absence of domain-specific prior knowledge and the highly camouflaged nature of objects can easily mislead SAM. While fine-tuning SAM with camouflage-specific data could mitigate these issues, the prohibitive computational costs and substantial data requirements render this approach impractical. Contemporary research \cite{ref12, ref14} has made strides in adapting SAM for COD applications, though existing works predominantly focus on fully supervised \cite{ref13, ref32} and weakly supervised \cite{ref4, ref9} paradigms. Notably, the exploration of SAM for SSCOD is still blank. Therefore, this work bridges this critical gap by synergistically combining self-training with SAM's strengths, injecting domain knowledge into SAM without fine-tuning, enabling its adaptation to complex COD tasks. Simultaneously harnessing SAM's superior segmentation capability to rectify noisy pseudo-labels, effectively addressing error accumulation in self-training. Finally, we achieves efficient, low-sample-dependent SSCOD through an elegantly concise architecture.

To address the aforementioned challenges in existing SSCOD methods, we proposes ST-SAM, a novel SAM-driven self-training framework for SSCOD. Our framework comprises two key components: Entropy-based Dynamic Filtering strategy and Domain Prompt-guided Mutual Correction strategy. Specifically, we first conduct preliminary training on a standard COD model using limited annotated data to acquire fundamental domain knowledge. Initial pseudo-labels generated from unlabeled data inevitably contain substantial noise due to task complexity and the model's early-stage limitations. To mitigate this, our entropy-based dynamic filtering strategy performs instance-level screening and ranking of pseudo-labels to identify relatively reliable predictions, preventing error propagation from challenging samples during early training phases. Subsequently, we compute pixel-level confidence scores for the filtered samples and implement weighted learning to minimize the impact of uncertain prediction regions. Concurrently, we leverage SAM's exceptional segmentation capability to address the severe noise in early-stage pseudo-labels. Inspired by SAM's interactive prompt-based segmentation, we convert the entropy-weighted pseudo-labels into specific hybrid prompts.  This innovative approach injects camouflage domain knowledge into SAM without requiring fine-tuning, enabling accurate segmentation of camouflaged targets. The framework then establishes mutual correction between SAM-generated pseudo-labels and entropy-weighted pseudo-labels to obtain high-confidence labels, effectively circumventing error accumulation. Through iterative expansion of the training set using these refined high-confidence pseudo-labels, the COD model progressively enhances its capabilities. Remarkably, ST-SAM achieves efficient end-to-end semi-supervised COD while requiring minimal labeled samples, and training of only a single network, significantly reducing computational overhead compared to existing multi-network approaches.

Our contributions can be summarized as follows:
\begin{compactitem}
	\item This paper for the first time investigate the potential of self-training strategies for SSCOD, constructing ST-SAM by combining the advantages of both self-training and SAM. This effectively resolves the strong dependence on annotation data in existing methods while maintaining strong scalability.
	\item We introduce EDF, which performs instance-level filtering and ranking of pseudo-labels, enabling the model to learn samples from easy to hard. Additionally, pixel-level weighting is applied to mitigate the negative impact of uncertain regions.
	\item We propose DPC, which converts weighted pseudo-labels into hybrid prompt information to inject domain knowledge into SAM. Leveraging SAM's powerful segmentation capability facilitates mutual correction of pseudo-labels.
	\item Extensive experiments results on 4 benchmark datasets demonstrate that ST-SAM achieves remarkable annotation efficiency, outperforming existing SSCOD methods with minimal labeled data (only 1\%) and even attaining performance comparable to fully supervised learning methods.
	
\end{compactitem}

\section{Related Work}

\subsection{Camouflaged Object Detection}

In recent years, with the emergence of open-source datasets \cite{ref18, ref20, ref19, ref21, ref40, ref41, ref50} and the rapid advancements in deep learning, significant progress has been made in COD research. For instance, SINet \cite{ref21} decomposes COD into a two-stage search-and-identification process. ZoomNet \cite{ref34} mimics the human observation mechanism for blurred images by proposing a hybrid-scale triplet network to handle scale variations in camouflaged objects. FRINet \cite{ref1} rethinks COD from a frequency-domain perspective, leveraging heterogeneous architectures to exploit spectral information. He et al. \cite{ref35} introduce text prompts as semantic cues for effective segmentation, while Hao et al. \cite{ref37} design a unified ViT framework for both COD and SOD with image reconstruction assistance.

Despite their impressive performance, existing methods share a critical limitation: heavy reliance on annotated data. Given the high annotation cost of COD datasets, large-scale data collection remains difficult, severely hindering further advancements in COD. To address this, weakly supervised and semi-supervised COD approaches have gained attention. He et al. \cite{ref2} propose a weakly supervised COD framework using scribble annotations to refine boundary details. Chen et al. \cite{ref3} construct the first weakly supervised COD dataset with point annotations. WS-SAM \cite{ref4} transforms weak labels into prompts for SAM \cite{ref5} to generate mask annotations. Niu et al. \cite{ref36} introduce a Mutual Interaction Network to mitigate ambiguity in scribble-based boundary information. Chen et al. \cite{ref9} further improve supervision by integrating multiple weak labels as prompts. In semi-supervised COD, research remains scarce. Lai et al. \cite{ref6} establish a baseline using consistency regularization to reduce pseudo-label noise, while Fu et al. \cite{ref7} propose an ensemble learning method to aggregate knowledge from different models and training stages. Additionally, Zhang et al. \cite{ref8} introduce Weakly-Semi-Supervised COD (WSSCOD), combining weak labels with partial manual annotations to enhance performance.

Semi-supervised COD holds great promise in alleviating annotation burdens while leveraging limited labeled data. However, current research is still in its early stages. To bridge this gap, we propose a more efficient semi-supervised COD framework that achieves superior performance while drastically reducing the demand for labeled samples.

\subsection{Application of SAM in COD}

The Segment Anything Model (SAM) has demonstrated remarkable performance and surprising zero-shot generalization capabilities in traditional segmentation tasks. However, recent studies \cite{ref10, ref11} have revealed that when directly applied to domain-specific tasks like Camouflaged Object Detection (COD), SAM suffers from significant performance degradation due to task complexity and domain gap caused by the lack of domain prior knowledge. While fine-tuning techniques could potentially adapt SAM to specific domains, they typically require massive computational resources and training samples, making them impractical for COD. To address these challenges, researchers have conducted a series of investigations. For the issue of SAM's lack of domain-specific knowledge, MCA-SAM \cite{ref14} and SAM-Adapter \cite{ref12} employ adapter techniques to reduce fine-tuning costs while enhancing SAM's domain-specific capabilities. TSP-SAM \cite{ref38} incorporates long-range spatiotemporal information into SAM for Video COD (VCOD). DSAM \cite{ref13} introduces depth modality information to improve SAM's performance in COD tasks. On another front, SAM's prompt-based interactive segmentation capability offers new opportunities for weakly supervised COD. Several studies have explored utilizing diverse weak supervision signals as prompts to leverage SAM's potential in COD tasks \cite{ref4, ref9}. However, the exploration of SAM's potential in semi-supervised COD remains completely unexplored.

In this work, we present the first investigation into SAM's potential for semi-supervised COD. We design a Domain Prompt-guided Mutual Correction strategy that can inject domain knowledge into SAM without requiring parameter fine-tuning, thereby enabling SAM to effectively learn from unlabeled samples.

\section{METHODOLOGY}
\subsection{Overview Architecture}

\begin{figure*}
	\begin{center}
		\includegraphics[scale=0.57]{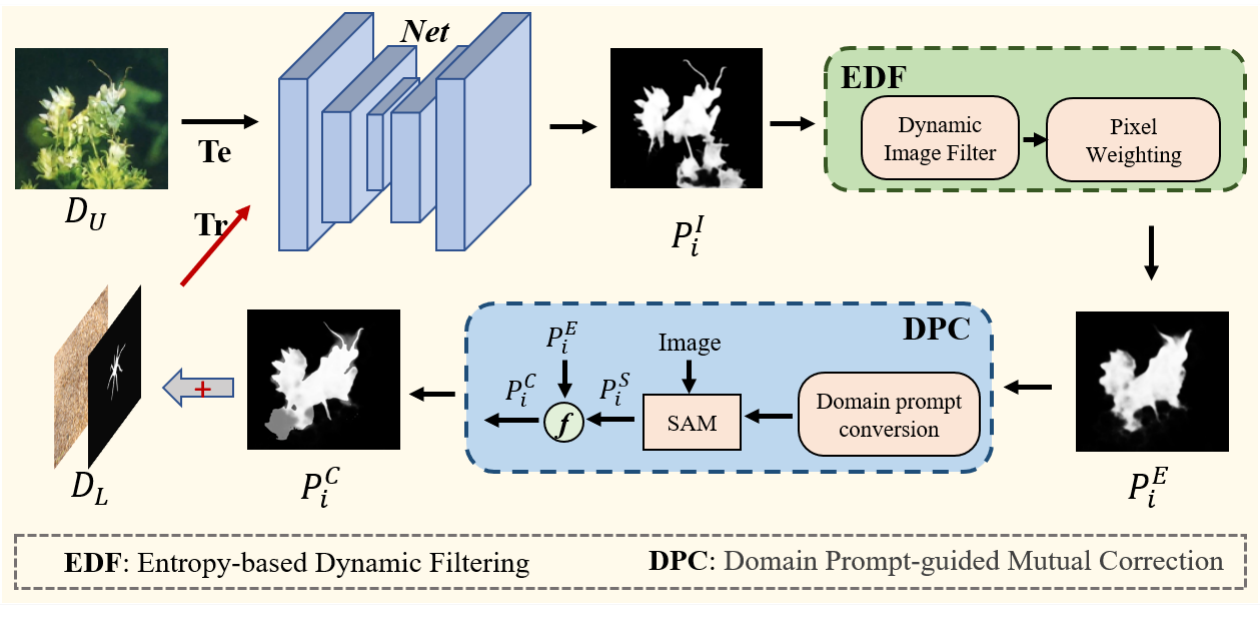}
	\end{center}
	\caption{The overall flowchart of ST-SAM. Tr: The Net is preliminarily trained using labeled data ${D_L}$; Te: The unlabeled data ${D_U}$ is processed through Net to obtain initial pseudo-labels $P_i^I$, which are then filtered using EDF to produce $P_i^E$. Finally, $P_i^E$ serves as domain prompts for DPC to make predictions and perform mutual correction, yielding high-confidence pseudo-labels $P_i^C$ for expansion.}
	\label{fig:2}
\end{figure*}

For semi-supervised camouflaged object detection (COD), the training set D consists of a labeled sample set ${D_L} = \left\{ {x_i^l,{y_i}} \right\}_{i = 1}^M$ and an unlabeled sample set ${D_U} = \left\{ {x_i^u} \right\}_{i = 1}^N$, where $x_i^{l/u}$ represents the input image and $y_i^{}$ denotes the corresponding mask label. To ensure higher annotation efficiency, usually $M<<N$.

Fig. \ref{fig:2} illustrates the overall architecture of proposed ST-SAM, which comprises three components: EDF, DPC, and a standard COD network (implemented as a basic encoder-decoder architecture \cite{ref8}, hereafter referred to as Net). The framework operates as follows: First, we pretrain Net on ${D_L}$ to acquire preliminary domain knowledge about camouflage patterns. The pretrained Net then processes ${D_U}$ to generate initial pseudo-labels $P_i^I$. Given the network's limited initial capability, these pseudo-labels contain numerous unreliable samples. We therefore apply EDF strategy to perform dual-level filtering: at the image level to select relatively reliable samples, followed by pixel-level weighting to filter out uncertain regions, ultimately producing entropy-weighted pseudo-labels $P_i^E$. This process is dynamic - the selection criteria gradually relax during training to incorporate more pseudo-labels to match the scale of the training set. In early training stages, both the scarcity of labeled data and network immaturity mean $P_i^E$ still lacks sufficient accuracy. To prevent error accumulation at this critical phase, we design the DPC that converts $P_i^E$ into domain-knowledge-enhanced hybrid prompts to guide SAM's segmentation. The SAM-generated masks $P_i^S$ are then compared and fused with $P_i^E$ to produce reliable co-corrected pseudo-labels $P_i^C$. Finally, we augment ${D_L}$ by incorporating $P_i^C$ with their corresponding images. This iterative process continues until all samples in ${D_U}$ are progressively incorporated into ${D_L}$, achieving effective utilization of unlabeled data. Through ST-SAM, we significantly reduce dependence on labeled data while achieving high-performance, efficient semi-supervised COD.

\subsection{Entropy-based Dynamic Filtering strategy}

Considering the difference in camouflage intensity among samples, we posit that a portion of the initial pseudo-labels $P_i^I$ is relatively reliable, with this proportion progressively increasing as training advances and model capability improves. Nevertheless, due to the limited ability of Net during early training stages, even relatively reliable samples inevitably contain uncertain regions. To address this, we design EDF to obtain more trustworthy pseudo-labels.

Specifically, for the initial pseudo-labels $P_i^I,i = 1,...,N$ obtained from the test set ${D_U}$, we first compute the local mean for each pixel using a window of size $\omega  \times \omega $:
\begin{equation}
	{p_f} = UF(Norm(P_i^I),\omega  \times \omega ).	
\end{equation}
Here, ${p_f}$ represents the local foreground probability, ${p_b} = 1 - {p_f}$, $Norm( \cdot )$ denotes normalization, and $UF( \cdot )$ represents mean filtering, $\omega  = 7$. Subsequently, the local entropy ${E_{local}}$ is computed to quantify the uncertainty in the neighborhood of each pixel:
\begin{equation}
	{E_{local}} =  - {p_f}\log ({p_f}) - {p_b}\log ({p_b}).	
\end{equation}
Similarly, we compute the global foreground probability $\mathop {{p_f}}\limits^{\thicksim }  = \frac{1}{{HW}}\sum {Norm(P_i^I)} $ for the entire mask to derive the global entropy ${E_{global}}$, which quantifies the uncertainty of the mask prediction:
\begin{equation}
	{E_{global}} =  - \mathop {{p_f}}\limits^{\thicksim } \log (\mathop {{p_f}}\limits^{\thicksim } ) - \mathop {{p_b}}\limits^{\thicksim } \log (\mathop {{p_b}}\limits^{\thicksim } ).	
\end{equation}
Subsequently, an uncertainty metric $u_\alpha$ is defined to assess the reliability of the pseudo-labels:
\begin{equation}
	{u_\alpha } = \frac{1}{N}\sum {\mathbb{I} } ({E_{local}} > {E_{global}} \times 0.5),	
\end{equation}
where $N$ is the total number of pixels, $\mathbb{I}( \cdot)$ is the indicator function. The sample retention condition is:
\begin{equation}
	Retain \; Sample \; If \; {u_\alpha } < {\tau _\alpha }, \; Else \; Discard.	
\end{equation}
where uncertainty threshold ${\tau _\alpha } = 0.3$.

Subsequently, for the retained masks, an entropy weight map is generated based on ${E_{local}}$ to weight them, thereby further mitigating the negative impact of uncertain regions. This process ultimately yields the entropy-weighted pseudo-labels $P_i^E$, as described below:
\begin{equation}
	P_i^E = P_i^I \cdot (0.5 + 0.5{(1 - {E_{local}})^k}).
\end{equation}
Here, the entropy weight decay coefficient $k=1$.

Finally, considering the limited performance of the Net in the early stages of training and the sample size $M <  < N$, it is necessary to dynamically expand the entropy-weighted pseudo-labels $P_i^E$. This allows the Net to learn samples from easy to difficult, progressively enhancing its ability to distinguish camouflaged objects and ultimately make accurate predictions for challenging samples. Specifically, the local entropy mean ${\mathop E\limits^ -  {_{local}}}$ of $P_i^E$ is computed to rank the samples. Assuming the current training set size is $x$, the top $x$ low-entropy samples $P_i^E,i = 1,...,x$ are selected as candidates for expansion. As the training epochs increase, $x$ will gradually increase, thereby achieving dynamic expansion.

\subsection{Domain Prompt-guided Mutual Correction strategy}

Due to the complex contours of camouflaged objects, obtaining accurate boundary predictions is highly challenging. Furthermore, limited by the scarcity of learning samples, the $P_i^E$ obtained in the early stages of training often fail to provide sufficient supervisory information, which represents another difficulty in self-training strategies. While SAM has demonstrated robust capabilities in segmentation tasks, it struggles to adapt to specific domains. Therefore, we designed DPC to effectively leverage the potential of SAM in weakly supervised self-training.

\begin{figure}
	\begin{center}
		\includegraphics[width=0.9\linewidth]{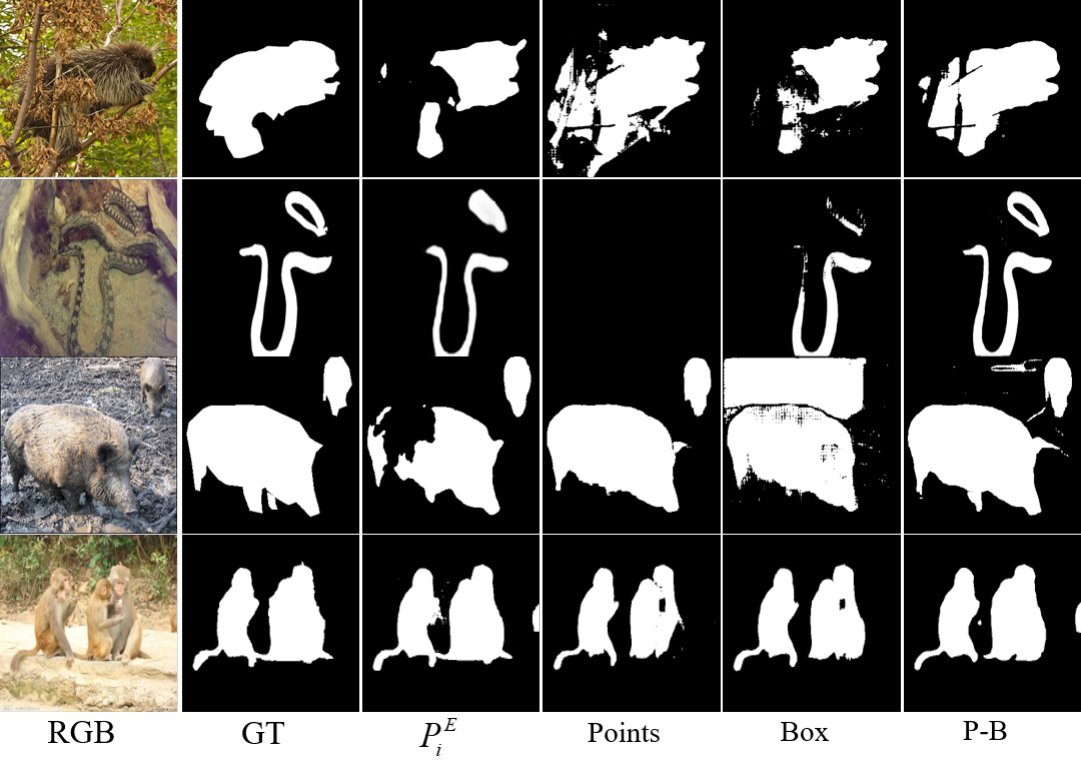}
	\end{center}
	\caption{Pseudo-labels generated by SAM based on different prompts.}
	\label{fig:3}
\end{figure}

Inspired by SAM's ability to perform interactive segmentation by incorporating user-provided prompts, we transform $P_i^E$ into domain-specific prompt to guide SAM in locating and segmenting camouflaged objects. Common prompt types supported by SAM include points, boxes, and scribbles. However, to avoid introducing additional manual annotations, we primarily consider points and boxes as prompts.

An intuitive approach is to sample the geometric center of each contour in $P_i^E$ to obtain multiple point prompts. This can effectively assist in locating multiple camouflaged objects. However, the reliability of this method depends on the completeness of $P_i^E$. As shown in the 1st row of Fig. \ref{fig:3}, when the quality of $P_i^E$ is poor and incomplete, the mask of a single object is divided into multiple disconnected regions, resulting in multiple point prompts. This misleads SAM into treating a single object as multiple objects, leading to incorrect segmentation. Similarly, generating multiple box prompts based on each region of the mask suffers from the same issue. Additionally, when the object is irregular, the sampled points may fall outside the target region, providing erroneous guidance and thus failing to achieve segmentation (2nd row of Fig. \ref{fig:3}). Calculating the minimum bounding rectangle for all regions of the mask as a single box prompt can maximize the integrity of the target. However, when multiple objects are sparsely distributed, the resulting prompt box may cover an excessively large area, thereby failing to provide effective localization guidance (3rd row of Fig. \ref{fig:3}).

Therefore, we design a hybrid prompt strategy to transform $P_i^E$ into reliable prompt information, effectively addressing the aforementioned issues. Specifically, for the multiple contour masks $Mask_i$ in $P^E$, we first filter out extremely small contour regions to prevent noise interference, resulting in $Mask_i^{'},i = 1,...,n$. Then, we extract the minimum bounding rectangle of $Mask_i^{'}$ as the box prompt $Prompt_B$:
\begin{equation}
	Promp{t_B} = Mask_i^{'}[{x_{\min }},{y_{\min }},{x_{\max }},{y_{\max }}].
\end{equation}

Next, we calculate the geometric center point ${c_i}({x_i},{y_i})$ of $Mask_i^{'}$ and verify whether ${c_i}({x_i},{y_i})$ lies inside $Mask_i^{'}$. If ${c_i}$ falls outside $Mask_i^{'}$, we employ an axial search strategy $AxialS( \cdot )$ to search along the positive and negative directions of the major axis from ${c_i}$ for the nearest point located inside $Mask_i^{'}$, which is designated as the safe center point ${c_i}^{'}= AxialS(c_i)$. This yields the point prompt set $Promp{t_P}$:
\begin{equation}
	Promp{t_P} = \left\{ {{c_i},IsInside({c_i}) = 1} \right\}\bigcup {\left\{ {c_i^{'},IsInside({c_i}) = 0} \right\}}.
\end{equation}

Then, $Promp{t_B}$ is combined with $Promp{t_P}$ to obtain the hybrid prompt $Promp{t_{P - B}}$. Through meticulously designed prompts, $P_i^E$ excels in providing domain-specific knowledge to guide SAM, even under conditions such as incomplete masks, multiple camouflaged objects, and irregular contours, resulting in the SAM pseudo-labels $P_i^S$. Finally, $P_i^E$ and $P_i^S$ are fused in equal proportion, leveraging the shared effective information to mutually correct potential errors in the pseudo-labels, thereby obtaining reliable corrected pseudo-labels $P_i^C$, which are dynamically expanded into the training set:
\begin{equation}
	P_i^C = fuse(P_i^E,P_i^S), P_i^S = SAM(Image,Promp{t_{P - B}}).
\end{equation}

\subsection{COD Model and Loss Function}

\begin{figure}
	\begin{center}
		\includegraphics[width=0.9\linewidth]{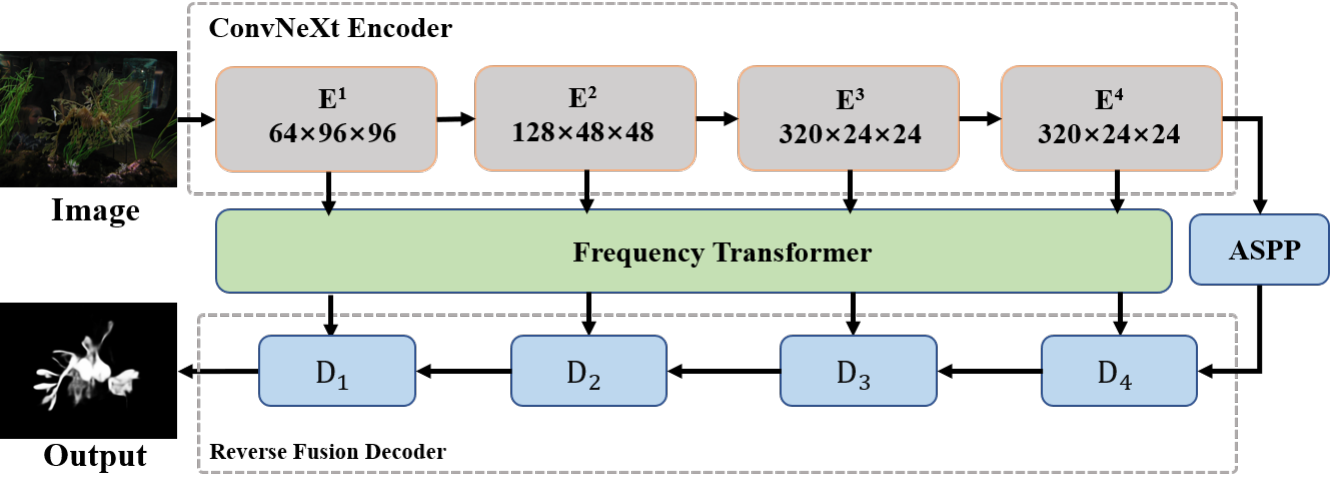}
	\end{center}
	\caption{The overall architecture of COD network, consisting of an encoder, Frequency Transformer, ASPP, and decoder.}
	\label{fig:4}
\end{figure}

The ST-SAM framework proposed in this paper is a flexible and scalable semi-supervised COD framework, in which the COD network is replaceable and does not constitute the primary contribution of this work. In our experiments, we adopted a classic U-Net COD network proposed in \cite{ref8}. As shown in Fig. \ref{fig:4}, the network employs ConvNeXt \cite{ref16} as the encoder to extract multi-scale features $\{ f_i^e\} _{i = 1}^4$ from the input. Subsequently, the Frequency Transformer $FT$ is utilized to capture fine details and deep semantics, yielding frequency-domain features $\{ f_i^F\} _{i = 1}^4$. For the high-level feature $f_4^e$, the ASPP \cite{ref17} module is applied to obtain deep semantic representation feature $f_4^E$. Then, in the reverse fusion decoder, the reverse mask block amplifies the differences between the background and foreground, enabling the convergence of multi-level features. Following the U-Net architecture, $f_4^E$ and $f_i^F$ are fused progressively to produce the camouflaged object prediction.

In this paper, a hybrid loss function is employed to supervise the predictions of the camouflaged object detection network. The total loss ${{\mathcal{L}}_{total}}$ consists of three components: the structural loss ${{\mathcal{L}}_S}$, the Dice loss ${{\mathcal{L}}_{dice}}$, and the uncertainty-aware loss ${{\mathcal{L}}_{UAL}}$. 
Specifically, ${{\mathcal{L}}_S} = wbce + wiou$ is used to balance the global structure and local details of the predictions. ${{\mathcal{L}}_{dice}}$ is employed to learn the boundary information of the target, defined as follows:
\begin{equation}
	{{\mathcal{L}}_{dice}} = 1 - \frac{{2 \cdot \sum {(Pre{d^p} \cdot Targe{t^p}) + S} }}{{\sum {Pre{d^p}}  + \sum { + Targe{t^p}}  + S}},
\end{equation}
where $p = 2$, and $S$ is a smoothing term. ${{\mathcal{L}}_{UAL}}$ is designed to address noise and ambiguous boundaries, enhancing the robustness of the model. It is defined as:
\begin{equation}
	{{\mathcal{L}}_{UAL}} = {\lambda _{ual}} \cdot \sum {(Pred \cdot \log (Pred) + (1 - Pred) \cdot \log (1 - Pred))},
\end{equation}
where ${\lambda _{ual}}$ is a dynamic weighting coefficient that decreases with the learning rate. The overall loss ${{\mathcal{L}}_{total}}$ of the model is defined as:
\begin{equation}
	{{\mathcal{L}}_{total}} = {{\mathcal{L}}_{s}} + \alpha  \cdot {{\mathcal{L}}_{dice}} + \beta  \cdot {{\mathcal{L}}_{UAL}},
\end{equation}
where $\alpha=4$, $\beta=2$.

\section{Experiment}

\begin{table*}[h]
	\centering
	\renewcommand{\tabcolsep}{1mm}
	\renewcommand{\arraystretch}{1}
	\scriptsize
	\caption{Quantitative comparison between our method and other 11 SOTA methods on 4 benchmark datasets. '-' indicates the code or result is not available.The optimal and suboptimal results are represented in $\textbf{\textcolor{red}{Red}}$ and $\textbf{\textcolor{blue}{Blue}}$.}
	\scalebox{1.1}{
		\begin{tabular}{c|ccccc|ccccc|ccccc|ccccc}
			\hline\toprule 
			\multirow{2}{*}{Methods} &\multicolumn{5}{c|}{CAMO} & \multicolumn{5}{c|}{CHAMELEON} & \multicolumn{5}{c|}{COD10K} & \multicolumn{5}{c}{NC4K} \\
             & $E_{\xi}\uparrow$ & $S_{\alpha}\uparrow$ & $F_\beta^\omega \uparrow$ & $F_{\beta}\uparrow$ & $\mathcal{M}\downarrow$ 
			& $E_{\xi}\uparrow$ & $S_{\alpha}\uparrow$ & $F_\beta^\omega \uparrow$ & $F_{\beta}\uparrow$ & $\mathcal{M}\downarrow$ 
			& $E_{\xi}\uparrow$ & $S_{\alpha}\uparrow$ & $F_\beta^\omega \uparrow$ & $F_{\beta}\uparrow$ & $\mathcal{M}\downarrow$ 
			& $E_{\xi}\uparrow$ & $S_{\alpha}\uparrow$ & $F_\beta^\omega \uparrow$ & $F_{\beta}\uparrow$ & $\mathcal{M}\downarrow$ \\
			\hline
			\rowcolor{blue!15}\multicolumn{21}{c}{\textbf{Fully-Supervised Methods}}	 \\
				
			\textbf{DSAM} \cite{ref13} \scalebox{0.7}{MM '24}
			& \textcolor{red}{0.906} & \textcolor{blue}{0.832} &  \textcolor{red}{0.794} & \textcolor{red}{0.824} & \textcolor{red}{0.061} 
			& - & - & - & - & - 
			& \textcolor{red}{0.913} & \textcolor{red}{0.846} & \textcolor{red}{0.760} & \textcolor{red}{0.761} & 0.033 
			& \textcolor{red}{0.928} & \textcolor{red}{0.871} & \textcolor{red}{0.826} & \textcolor{red}{0.845} & \textcolor{red}{0.040} \\
			
			\textbf{TJNet} \cite{ref25} \scalebox{0.7}{AI '24}
			& \textcolor{blue}{0.890} &  \textcolor{red}{0.841} & \textcolor{blue}{0.779} & - & \textcolor{blue}{0.064} 
			& \textcolor{red}{0.958} &  \textcolor{red}{0.913} &  \textcolor{red}{0.859} & - & \textcolor{red}{0.024} 
			& \textcolor{blue}{0.907} & \textcolor{blue}{0.844} & \textcolor{blue}{0.738} & - & \textcolor{red}{0.030} 
			& - & - & - & - & - \\
            
      \textbf{ICEG} \cite{ref27} \scalebox{0.7}{ICLR '24}
			& 0.879 & 0.810 & - & 0.789 &  0.068
			& \textcolor{blue}{0.950} & \textcolor{blue}{0.899} & - & \textcolor{red}{0.858} & \textcolor{blue}{0.027}
			& 0.906 & 0.826 & - & \textcolor{blue}{0.747} & \textcolor{red}{0.030} 
			& 0.908 & 0.849 & - & 0.814 & 0.044 \\
			
			\textbf{DINet} \cite{ref26} \scalebox{0.7}{TMM '24}
			&  0.883 & 0.821 & - &  \textcolor{blue}{0.790} &  0.068
			& - & - & - & - & - 
			& 0.901 & 0.832 & - & 0.744 & \textcolor{blue}{0.031} 
			& \textcolor{blue}{0.910} & \textcolor{blue}{0.856} & - & \textcolor{blue}{0.820} & \textcolor{blue}{0.043} \\
			
			\rowcolor{blue!15}\multicolumn{21}{c}{\textbf{Weakly-Supervised Methods}}	 \\
			
			\textbf{WS-SAM} \cite{ref4} \scalebox{0.7}{NeurIPS '23} 
			& 0.818 & 0.759 & - & \textcolor{red}{0.742} & 0.092
			& \textcolor{blue}{0.897} & \textcolor{blue}{0.824} & - & \textcolor{blue}{0.777} & \textcolor{blue}{0.046} 
			& \textcolor{red}{0.878} & \textcolor{blue}{0.803} & - & \textcolor{red}{0.719} & \textcolor{red}{0.038} 
			& \textcolor{blue}{0.886} & \textcolor{red}{0.829} & - & \textcolor{red}{0.802} & \textcolor{blue}{0.052} \\
			
			\textbf{PSCOD} \cite{ref3} \scalebox{0.7}{ECCV '24}
			& \textcolor{red}{0.872} & \textcolor{red}{0.798} & \textcolor{red}{0.727} & - & \textcolor{red}{0.074} 
			& - & - & - & - & - 
			& 0.859 & 0.784 & \textcolor{red}{0.650} & - & \textcolor{blue}{0.042} 
			& \textcolor{red}{0.889} & \textcolor{blue}{0.822} & \textcolor{red}{0.748} & - & \textcolor{red}{0.051} \\

			\textbf{CRNet} \cite{ref2} \scalebox{0.7}{AAAI '23} 
			& 0.815 & 0.735 & \textcolor{blue}{0.641} & - & 0.092
			& \textcolor{blue}{0.897} & 0.818 & \textcolor{red}{0.744} & - & \textcolor{blue}{0.046} 
			& 0.832 & 0.733 & \textcolor{blue}{0.576} & - & 0.049
			& - & - & - & - & - \\

			\textbf{ProMaC} \cite{ref23} \scalebox{0.7}{NeurIPS '24} 
			& \textcolor{blue}{0.846} & \textcolor{blue}{0.767} & - & \textcolor{blue}{0.725} & \textcolor{blue}{0.090} 
			& \textcolor{red}{0.899} & \textcolor{red}{0.833} & - & \textcolor{red}{0.790} & \textcolor{red}{0.044} 
			& \textcolor{blue}{0.876} & \textcolor{red}{0.805} & - & \textcolor{blue}{0.716} & \textcolor{blue}{0.042} 
			& - & - & - & - & - \\ 
      
      \textbf{GenSAM} \cite{ref24} \scalebox{0.7}{AAAI '24} 
			& 0.775 & 0.719 & - & 0.659 & 0.113 
			& 0.807 & 0.764 & - & 0.680 & 0.090 
			& 0.838 & 0.775 & - & 0.681 & 0.067 
			& - & - & - & - & - \\ 
			
            \bottomrule
			\rowcolor{blue!15}\multicolumn{21}{c}{\textbf{Semi-Supervised Methods}}	 \\
            
      \textbf{CamoTeacher} \cite{ref6} -1\% 
			& 0.669 & 0.621 & 0.456 & 0.545 & 0.136 
			& 0.714 & 0.652 & 0.476 & 0.558 & 0.093 
			& 0.788 & 0.699 & 0.517 & 0.582 & 0.062 
			& 0.779 & 0.718 & 0.599 & 0.675 & 0.090 \\
			
			\textbf{CamoTeacher} -5\% 
			& 0.711 & 0.669 & 0.523 & 0.601 & 0.122 
			& 0.785 & 0.729 & 0.587 & 0.656 & 0.070 
			& 0.827 & 0.745 & 0.583 & 0.644 & 0.050 
			& 0.834 & 0.777 & 0.677 & 0.739 & 0.071 \\
			
			\textbf{CamoTeacher} -10\% \scalebox{0.7}{ECCV '24}
			& 0.742 & 0.701 & 0.560 & 0.635 & 0.112 
			& 0.813 & 0.756 & 0.617 & 0.684 & 0.065 
			& 0.836 & 0.759 & 0.594 & 0.652 & 0.049 
			& 0.842 & 0.791 & 0.687 & 0.746 & 0.068 \\
            
            \textbf{SSCOD} \cite{ref7} -1\% 
			& 0.804 & 0.708 & 0.583 & 0.653 & 0.110 
			& 0.771 & 0.683 & 0.574 & 0.629 & 0.063 
			& 0.805 & 0.725 & 0.537 & 0.578 & 0.057 
			& 0.844 & 0.767 & 0.652 & 0.700 & 0.073 \\
			
			\textbf{SSCOD} -10\% 
			& 0.806 & 0.737 & 0.638 & 0.708 & 0.094 
			& 0.878 & 0.805 & 0.707 & 0.751 & 0.047 
			& 0.852 & 0.779 & 0.639 & 0.676 & 0.042 
			& 0.868 & 0.808 & 0.729 & 0.775 & 0.059 \\
			
			\textbf{SSCOD} -20\% \scalebox{0.7}{MM '24}
			& 0.844 & 0.778 & 0.704 & 0.767 & 0.078 
			& 0.906 & 0.834 & 0.761 & 0.802 & 0.042 
			& 0.864 & 0.791 & 0.662 & 0.699 & 0.039 
			& 0.882 & 0.821 & 0.750 & 0.795 & 0.055 \\
             
			\bottomrule
			\textbf{Ours} -1\% 
			& 0.879 & 0.819 & 0.753 & 0.804 & 0.070 
			& 0.920 & 0.848 & \textcolor{blue}{0.778} & \textcolor{blue}{0.815} & 0.038 
			& \textcolor{red}{0.884} & 0.824 & \textcolor{blue}{0.713} & \textcolor{red}{0.737} & 0.032 
			& 0.907 & 0.855 & 0.795 &\textcolor{red}{0.830} & 0.043 \\
			
			\textbf{Ours} -5\%
			& \textcolor{blue}{0.886} & 0.831 & 0.761 & 0.807 & 0.066 
			& \textcolor{blue}{0.921} & \textcolor{blue}{0.855} & 0.776 & 0.798 & \textcolor{blue}{0.036} 
			& 0.871 & \textcolor{blue}{0.834} & 0.709 & 0.716 & 0.031 
			& \textcolor{blue}{0.909} & \textcolor{blue}{0.870} & \textcolor{blue}{0.803} & 0.823 & \textcolor{blue}{0.038} \\
			
			\textbf{Ours} -10\% 
			& \textcolor{blue}{0.886} & \textcolor{blue}{0.835} & \textcolor{blue}{0.778} & \textcolor{red}{0.818} & \textcolor{blue}{0.063} 
			& 0.919 & 0.850 & 0.770 & 0.792 & \textcolor{blue}{0.036} 
			& \textcolor{blue}{0.882} & \textcolor{red}{0.837} & \textcolor{red}{0.723} & \textcolor{blue}{0.729} & \textcolor{red}{0.029} 
			& \textcolor{red}{0.911} & 0.868 & \textcolor{red}{0.807} & \textcolor{red}{0.830} & \textcolor{blue}{0.038} \\
			
			\textbf{Ours} -20\% 
			& \textcolor{red}{0.890} & \textcolor{red}{0.844} &\textcolor{red}{0.779} & \textcolor{blue}{0.809} & \textcolor{red}{0.058} 
			& \textcolor{red}{0.926} & \textcolor{red}{0.876} & \textcolor{red}{0.804} & \textcolor{red}{0.818} & \textcolor{red}{0.032} 
			& 0.874 & \textcolor{red}{0.837} & \textcolor{blue}{0.713} & 0.717 & \textcolor{blue}{0.030} 
			& \textcolor{red}{0.911} & \textcolor{red}{0.874} & \textcolor{red}{0.807} & \textcolor{blue}{0.824} & \textcolor{red}{0.037} \\
			
			\bottomrule
			\hline
	\end{tabular}}
	\label{tab:1}
\end{table*}

\subsection{Experimental Setup}
\textbf{Datasets.} In this paper, the proposed ST-SAM is evaluated on 4 COD benchmark datasets: CAMO \cite{ref18}, CHAMELEON \cite{ref19}, COD10K \cite{ref20}, and NC4K \cite{ref21}. CAMO consists of 1,250 challenging camouflaged images, with 1,000 images for training and 250 images for testing. CHAMELEON contains 76 images of camouflaged animals. COD10K includes 5,066 camouflaged images, with 3,040 images for training and 2,026 images for testing. NC4K comprises 4,121 camouflaged images collected from the web. Following the same dataset partitioning as in other studies \cite{ref6, ref7}, we construct the training set using 1,000 samples from CAMO and 3,040 samples from COD10K. Then, the entire NC4K and CHAMELEON datasets, along with the remaining samples from CAMO and COD10K, are used as the test set to evaluate the performance of ST-SAM and competing models. For the training set, we adhere to the semi-supervised learning partitioning strategy \cite{ref6, ref7}, randomly sampling 1\%, 5\%, 10\%, and 20\% of the labeled data from the training set as the labeled sample set ${D_L}$, with the remaining portion serving as the unlabeled sample set ${D_U}$.

\textbf{Evaluation Metrics.} Following previous works \cite{ref6, ref7}, we adopt five widely used evaluation metrics to assess the performance of our model. These include E-measure (${E_\xi }$) \cite{ref42}, S-measure (${S_\alpha }$) \cite{ref43}, Weighted F-score ($F_\beta ^\omega $) \cite{ref28}, F-measure (${F_\beta }$) \cite{ref44}, and Mean Absolute Error (MAE). Here, ${E_\xi }$ and ${F_\beta }$ represent adaptive values.

\textbf{Implementation Details.} To enhance the robustness, we employ data augmentation strategies such as random flipping, rotation, and boundary cropping on the training images to prevent overfitting. During both the training and testing phases, the input image size is resized to 384$\times$384. The COD network initializes the encoder parameters using ConvNeXt-B \cite{ref16}, while SAM \cite{ref5} adopts the parameters and settings of vit-h. We utilize the Adam optimizer \cite{ref22} to train our network, with a batch size set to 8. The initial learning rate is set to 1e-7, linearly warmed up to 1e-4 within 10 epochs, and then cosine annealed to 1e-7 over 150 epochs to complete the preliminary training of the COD network. Subsequently, the learning rate is reset to 1e-4 and dynamically expanded and trained using the proposed two strategies every 20 epochs, with cosine annealing to 1e-7 over 300 epochs to complete the training.

\subsection{Comparison with state-of-the-art methods}
To validate the effectiveness of ST-SAM, we compared it with 11 SOTA methods, including semi-supervised methods: CamoTeacher \cite{ref6}, SSCOD \cite{ref7}; weakly supervised methods: WS-SAM \cite{ref4}, PSCOD \cite{ref3}, CRNet \cite{ref2}, ProMaC \cite{ref23}, GenSAM \cite{ref24}; and fully supervised methods: DSAM \cite{ref13}, TJNet \cite{ref25}, DINet \cite{ref26}, ICEG \cite{ref27}. To ensure a fair comparison, the results of these methods were either obtained from publicly available data or generated by training their source code. The comparison results are as follows: 

\textbf{Quantitative Comparison.} The quantitative comparison results of the proposed algorithm and the 11 other methods on the four datasets are shown in Table \ref{tab:1}. From the results, it can be observed that, compared to the existing two semi-supervised methods, ST-SAM achieves the best performance under the same amount of labeled samples. Additionally, under the condition of scarce labeled samples (1\%), the performance of the other two algorithms significantly declines, while ST-SAM maintains excellent performance, even outperforming the versions of existing methods with 20\% labeled samples. This demonstrates that ST-SAM effectively addresses the strong dependency on labeled samples in semi-supervised COD. Furthermore, we compared ST-SAM with weakly supervised and fully supervised methods. It can be seen that ST-SAM outperforms SOTA weakly supervised COD methods and even achieves competitive performance compared to fully supervised methods. These experiments validate the effectiveness of ST-SAM and highlight the great potential of semi-supervised COD strategies.

\begin{figure*}
	\begin{center}
		\includegraphics[scale=0.48]{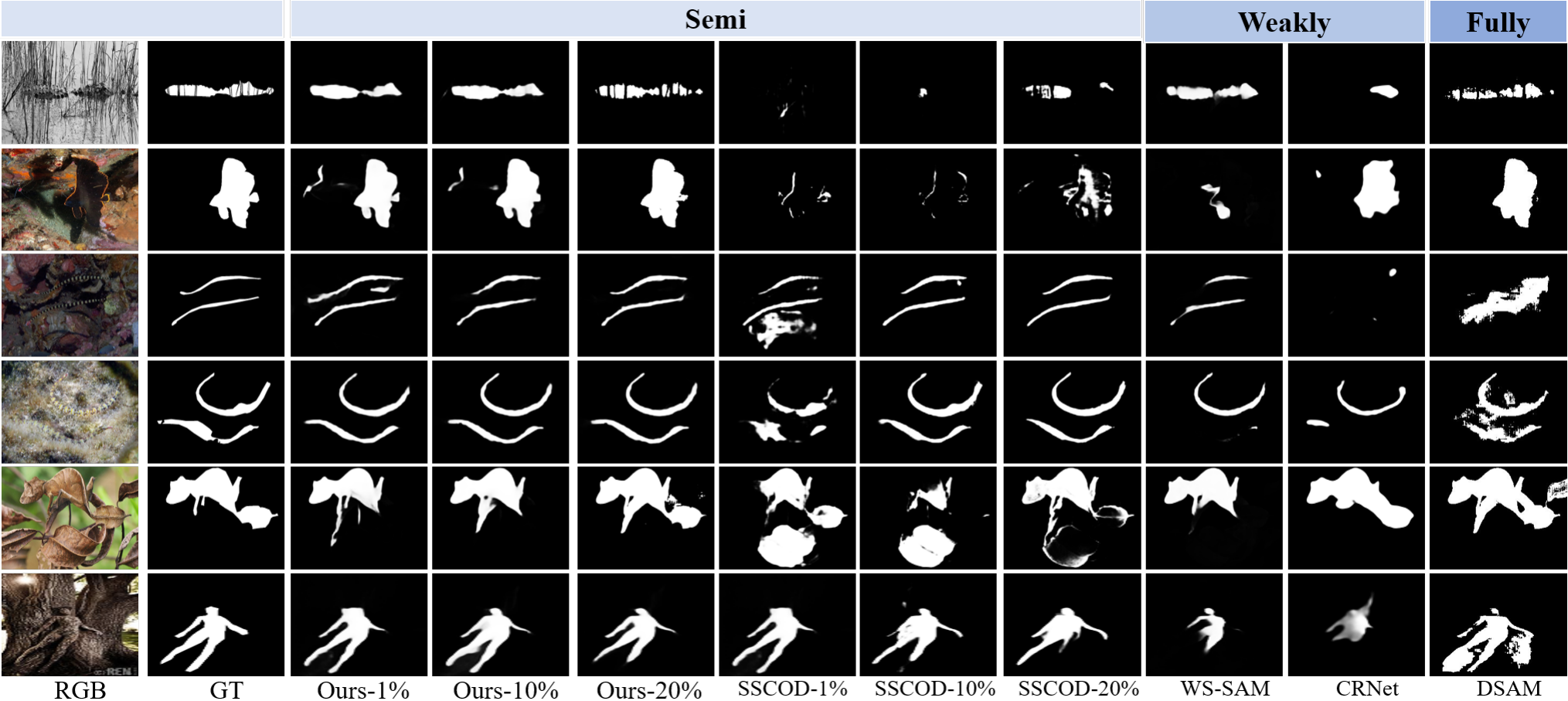}
	\end{center}
	\caption{Qualitative comparison results between ST-SAM and SOTA COD models.}
	\label{fig:5}
\end{figure*}

\textbf{Qualitative Comparison.} To visually demonstrate the capabilities of our method, Fig. \ref{fig:5} presents a visual comparison of ST-SAM with other semi-supervised, weakly supervised, and fully supervised SOTA algorithms in some representative camouflaged scenarios. From the results, it is evident that ST-SAM, with only 1\% labeled samples, can effectively detect and segment camouflaged objects in most scenarios, even outperforming semi-supervised methods with 20\% labeled samples and weakly supervised methods. When the labeled samples are increased to 20\%, the performance of ST-SAM further improves, enabling it to more effectively handle complex scenarios such as occlusions (Row 1) and small objects (Rows 3, 4), significantly reducing misjudgments (Row 2) and producing more complete segmentation results (Rows 5, 6). ST-SAM can match or even surpass fully supervised methods.

\subsection{Ablation Study}
We conducted experiments to demonstrate the effectiveness of ST-SAM. All experiments were performed with 1\% labeled samples.
Due to space limitations, the analysis of complexity, failure cases, ablation of hyper-parameters, and some details are provided in supplementary materials.
\begin{table}[h]
	\centering
	\renewcommand{\tabcolsep}{1.5mm}
	\renewcommand{\arraystretch}{0.8}
	\scriptsize
	\caption{Ablation experiments on the effectiveness of each component, the best results are marked in $\textbf{\textcolor{red}{Red}}$.} 
	\scalebox{1}{
		\begin{tabular}{c|c|cHccc|cHcccHHHHH}
			\hline\toprule	
      \multirow{2}{*}{\centering ID}&\multirow{2}{*}{\centering Variants}&\multicolumn{5}{c|}{\centering COD10K} & \multicolumn{5}{c}{\centering NC4K} \\
			&\multicolumn{1}{c|}{}
			&$E_{\xi}\uparrow$ &$S_{\alpha}\uparrow$ &$F_\beta^\omega \uparrow$ &$F_{\beta}\uparrow$ &$\mathcal{M}\downarrow$
			&$E_{\xi}\uparrow$ &$S_{\alpha}\uparrow$ &$F_\beta^\omega \uparrow$ &$F_{\beta}\uparrow$ &$\mathcal{M}\downarrow$
			&$E_{\xi}\uparrow$ &$S_{\alpha}\uparrow$ &$F_\beta^\omega \uparrow$ &$F_{\beta}\uparrow$ &$\mathcal{M}\downarrow$ \\
			\midrule		
			1 & Baseline & 0.793 & 0.732 & 0.517 & 0.569 & 0.054 & 0.834 & 0.760 & 0.611 & 0.693 & 0.077 & 0.783 & 0.664 & 0.493 & 0.632 & 0.126 \\
			2 & Self-Training & 0.807 & 0.718 & 0.535 & 0.585 & 0.057 & 0.830 & 0.748 & 0.628 & 0.688 & 0.078 & 0.749 & 0.653 & 0.509 & 0.611 & 0.127 \\
			3 & +EDF & 0.868 & 0.797 & 0.676 & 0.708 & 0.037 & 0.883 & 0.816 & 0.748 & 0.804 & 0.056  & 0.806 & 0.728 & 0.639 & 0.742 & 0.099 \\
			4 & +DPC & 0.856 & 0.794 & 0.655 & 0.684 & 0.038 & 0.879 & 0.826 & 0.740 & 0.778 & 0.052 & 0.836 & 0.768 & 0.672 & 0.733 & 0.088 \\  
			5 & +EDF+DPC & \textcolor{red}{0.884} & \textcolor{red}{0.824} & \textcolor{red}{0.713} & \textcolor{red}{0.737} & \textcolor{red}{0.032} 
			& \textcolor{red}{0.907} & \textcolor{red}{0.855} & \textcolor{red}{0.795} &\textcolor{red}{0.830} & \textcolor{red}{0.043} & \textcolor{red}{0.879} & \textcolor{red}{0.819} & \textcolor{red}{0.753} & \textcolor{red}{0.804} & \textcolor{red}{0.070}\\ 
			
			\bottomrule
			\hline
	\end{tabular}}
	\label{tab:2}
\end{table}

\textbf{Ablation on EDF and DPC.} To validate the effectiveness of the proposed EDF and DPC, the following experiments were conducted: (1) Learning only ${D_L}$ as the baseline; (2) A classic self-training strategy, where pseudo-labels are obtained from ${D_U}$ using the COD network to directly expand ${D_L}$; (3) Incorporating the EDF strategy into the COD network; (4) Incorporating the DPC strategy into the COD network; (5) Combining both the EDF and DPC strategies in the COD network, i.e., the ST-SAM framework.

Taking $F_\beta ^\omega $ of COD10K as an example, Table \ref{tab:2} shows that due to the complexity of camouflaged scenes, directly applying the self-training strategy fails to unlock the potential of ${D_U}$, providing almost no performance gain. The EDF strategy reduces error accumulation through global and local filtering as well as dynamic expansion, improving performance by 15.9\% compared to the baseline. The DPC strategy compensates for the shortcomings of the initial model by leveraging domain knowledge to drive SAM, improving performance by 13.8\% compared to the baseline, thus proving their effectiveness. Furthermore, the combination of both strategies further enhances performance to 19.6\%, demonstrating the rationality of the ST-SAM framework design.

\begin{table}[h]
	\centering
	\renewcommand{\tabcolsep}{1.5mm}
	\renewcommand{\arraystretch}{0.8}
	\scriptsize
	\caption{Ablation experiments on the effectiveness of EDF, the results marked in $\textbf{\textcolor{red}{Red}}$ indicate the best performance.} 
	\scalebox{1}{
		\begin{tabular}{c|c|cHccc|cHcccHHHHH}
			\hline\toprule
			
			\multirow{2}{*}{\centering ID}&\multirow{2}{*}{\centering Variants}&\multicolumn{5}{c|}{\centering COD10K} & \multicolumn{5}{c}{\centering NC4K} \\
			&\multicolumn{1}{c|}{}
			&$E_{\xi}\uparrow$ &$S_{\alpha}\uparrow$ &$F_\beta^\omega \uparrow$ &$F_{\beta}\uparrow$ &$\mathcal{M}\downarrow$
			&$E_{\xi}\uparrow$ &$S_{\alpha}\uparrow$ &$F_\beta^\omega \uparrow$ &$F_{\beta}\uparrow$ &$\mathcal{M}\downarrow$
			&$E_{\xi}\uparrow$ &$S_{\alpha}\uparrow$ &$F_\beta^\omega \uparrow$ &$F_{\beta}\uparrow$ &$\mathcal{M}\downarrow$ \\
			\midrule

			1 & One-shot & 0.817 & 0.779 & 0.621 & 0.649 & 0.047 & 0.872 & 0.826 & 0.737 & 0.775 & 0.055 & 0.839 & 0.770 & 0.676 & 0.736 & 0.091 \\
			2 & Equal Ratio & 0.850 & 0.812 & 0.682 & 0.703 & 0.036 & 0.891 & 0.848 & 0.778 & 0.812 & 0.045 & 0.877 & \textcolor{red}{0.821} & \textcolor{red}{0.753} & 0.802 & \textcolor{red}{0.069} \\
			3 & Epoch-Dynamic & \textcolor{red}{0.884} & \textcolor{red}{0.824} & \textcolor{red}{0.713} & \textcolor{red}{0.737} & \textcolor{red}{0.032} 
			& \textcolor{red}{0.907} & \textcolor{red}{0.855} & \textcolor{red}{0.795} &\textcolor{red}{0.830} & \textcolor{red}{0.043} & \textcolor{red}{0.879} & 0.819 & \textcolor{red}{0.753} & \textcolor{red}{0.804} & 0.070 \\  \midrule
			4 & H→L Entropy &  0.788 & 0.722 & 0.525 & 0.590 & 0.059 & 0.813 & 0.751 & 0.612 & 0.685 & 0.078 & 0.787 & 0.698 & 0.549 & 0.650 & 0.109 \\ 
			5 & Random Select & 0.883 & 0.820 & 0.710 & \textcolor{red}{0.737} & \textcolor{red}{0.032} & 0.902 & 0.850 & 0.789 & \textcolor{red}{0.830} & 0.044 & 0.864 & 0.792 & 0.721 & 0.794 & 0.078 \\ 
			6 & L→H Entropy & \textcolor{red}{0.884} & \textcolor{red}{0.824} & \textcolor{red}{0.713} & \textcolor{red}{0.737} & \textcolor{red}{0.032} 
			& \textcolor{red}{0.907} & \textcolor{red}{0.855} & \textcolor{red}{0.795} &\textcolor{red}{0.830} & \textcolor{red}{0.043} & \textcolor{red}{0.879} & \textcolor{red}{0.819} & \textcolor{red}{0.753} & \textcolor{red}{0.804} & \textcolor{red}{0.070} \\ 
			
			\bottomrule
			\hline
	\end{tabular}}
	\vspace{-\baselineskip}
	\label{tab:3}
\end{table}

\textbf{Ablation of EDF.} We conducted the following experiments to validate the rationality of the EDF, with the results shown in Table \ref{tab:3}, using $F_\beta ^\omega $ of COD10K as an example for comparison.

Regarding the dynamic expansion strategy for pseudo-labels: (1) Expanding all qualified samples at once for learning; (2) Expanding an equal proportion (20\%) of samples at each step for learning; (3) Dynamically expanding the proportion of samples as training epochs progress. From the results, it can be observed that due to the limited ability of the early model to learn difficult samples, expanding all samples at once yields the worst performance at only 62.1\%. Expanding an equal proportion effectively mitigates this issue, improving performance by 6.1\%. The dynamic expansion strategy gradually increases the proportion of samples as the model’s capability improves, achieving a 9.2\% improvement.

Regarding the sample learning order: (4) Learning high-entropy samples first, followed by low-entropy samples; (5) Randomly selecting samples from the qualified candidates; (6) Learning low-entropy samples first, followed by high-entropy samples. From the results, it can be seen that although for many semi-supervised self-training tasks, prioritizing high-entropy samples can better learn complex patterns and enhance generalization, making it a more common choice, for COD tasks, the complexity of camouflaged scenes leads to performance collapse (only 52.5\%) when learning high-entropy samples too early without sufficient reliable information, which is far lower than random selection. Gradually enhancing the model by expanding from low-entropy to high-entropy samples brings further improvement compared to random selection.

\begin{table}[h]
	\centering
	\renewcommand{\tabcolsep}{1.5mm}
	\renewcommand{\arraystretch}{0.8}
	\scriptsize
	\caption{Ablation experiments on the effectiveness of DPC, the results marked in $\textbf{\textcolor{red}{Red}}$ indicate the best performance.} 
	\scalebox{1}{
		\begin{tabular}{c|c|cHccc|cHcccHHHHH}
			\hline\toprule
			
			\multirow{2}{*}{\centering ID}&\multirow{2}{*}{\centering Variants}&\multicolumn{5}{c|}{\centering COD10K} & \multicolumn{5}{c}{\centering NC4K} \\
			&\multicolumn{1}{c|}{}
			&$E_{\xi}\uparrow$ &$S_{\alpha}\uparrow$ &$F_\beta^\omega \uparrow$ &$F_{\beta}\uparrow$ &$\mathcal{M}\downarrow$
			&$E_{\xi}\uparrow$ &$S_{\alpha}\uparrow$ &$F_\beta^\omega \uparrow$ &$F_{\beta}\uparrow$ &$\mathcal{M}\downarrow$
			&$E_{\xi}\uparrow$ &$S_{\alpha}\uparrow$ &$F_\beta^\omega \uparrow$ &$F_{\beta}\uparrow$ &$\mathcal{M}\downarrow$ \\
			\midrule
			
			1 & Points & 0.830 & 0.787 & 0.645 & 0.681 & 0.050 & 0.872 & 0.824 & 0.742 & 0.794 & 0.061 & 0.840 & 0.781 & 0.694 & 0.764 & 0.085 \\
			2 & Box & 0.876 & 0.821 & 0.701 & 0.731 & 0.035 & 0.905 & 0.852 & \textcolor{red}{0.795} & 0.827 & 0.044 & 0.866 & 0.796 & 0.730 & 0.800 & 0.077 \\
			3 & P-B & \textcolor{red}{0.884} & \textcolor{red}{0.824} & \textcolor{red}{0.713} & \textcolor{red}{0.737} & \textcolor{red}{0.032} 
			& \textcolor{red}{0.907} & \textcolor{red}{0.855} & \textcolor{red}{0.795} &\textcolor{red}{0.830} & \textcolor{red}{0.043} & \textcolor{red}{0.879} & \textcolor{red}{0.819} & \textcolor{red}{0.753} & \textcolor{red}{0.804} & \textcolor{red}{0.070} \\  \midrule
			4 & Intersect & 0.821 & 0.745 & 0.581 & 0.649 & 0.049 & 0.826 & 0.752 & 0.639 & 0.740 & 0.077 & 0.782 & 0.677 & 0.544 & 0.692 & 0.117 \\ 
			5 & Union & 0.800 & 0.779 & 0.622 & 0.626 & 0.048 & 0.871 & 0.838 & 0.752 & 0.762 & 0.049 & 0.866 & 0.815 & 0.737 & 0.766 & 0.072 \\ 
			6 & Ratio & \textcolor{red}{0.884} & \textcolor{red}{0.824} & \textcolor{red}{0.713} & \textcolor{red}{0.737} & \textcolor{red}{0.032} 
			& \textcolor{red}{0.907} & \textcolor{red}{0.855} & \textcolor{red}{0.795} &\textcolor{red}{0.830} & \textcolor{red}{0.043} & \textcolor{red}{0.879} & \textcolor{red}{0.819} & \textcolor{red}{0.753} & \textcolor{red}{0.804} & \textcolor{red}{0.070} \\ 
			\bottomrule
			\hline
	\end{tabular}}
	\label{tab:4}
\end{table}

\textbf{Ablation of DPC.} We conducted the following experiments to validate the rationality of the DPC, with the results shown in Table \ref{tab:4}, using $F_\beta ^\omega $ of COD10K as an example for comparison.

Regarding different prompt types: (1) Multiple point prompts; (2) A single box prompt; (3) A hybrid prompt combining multiple points and a single box. From the results, it can be observed that due to the complex morphology of camouflaged objects, converting limited-quality pseudo-labels into point prompts provides misleading guidance, resulting in incomplete object segmentation and poor performance at only 64.5\%. In contrast, the box prompt, leveraging region information, effectively complements missing areas, achieving a result of 70.1\%. The hybrid prompt adopted in this paper combines the advantages of both, effectively addressing scenarios with multiple objects and incomplete regions, further improving performance to 71.3\%.

Regarding the fusion method of $P_i^E$ and $P_i^S$: (4) Taking the intersection of the two labels; (5) Taking the union of the two labels; (6) Fusing the two labels in equal proportion. From the results, it can be seen that taking the intersection of pseudo-labels filters out erroneous regions but leads to false negatives. For complex COD scenes, the lack of sufficient guidance can easily cause performance collapse, achieving only 58.1\%. Taking the union of pseudo-labels ensures completeness but introduces false positives, bringing in many errors, achieving 62.2\%. In contrast, proportionally fusing the pseudo-labels maximizes information retention while suppressing potential errors, achieving the optimal performance of 71.3\%.

\begin{table}[h]
	\centering
	\renewcommand{\tabcolsep}{1.2mm}
	\renewcommand{\arraystretch}{1}
	\scriptsize
	\caption{Validation of the scalability of ST-SAM, the results marked in $\textbf{\textcolor{red}{Red}}$ indicate the best performance.} 
	\scalebox{1}{
		\begin{tabular}{c|c|cHccc|cHcccHHHHH}
			\hline\toprule
			
			\multirow{2}{*}{\centering ID}&\multirow{2}{*}{\centering Variants}&\multicolumn{5}{c|}{\centering COD10K} & \multicolumn{5}{c}{\centering NC4K} \\
			&\multicolumn{1}{c|}{}
			&$E_{\xi}\uparrow$ &$S_{\alpha}\uparrow$ &$F_\beta^\omega \uparrow$ &$F_{\beta}\uparrow$ &$\mathcal{M}\downarrow$
			&$E_{\xi}\uparrow$ &$S_{\alpha}\uparrow$ &$F_\beta^\omega \uparrow$ &$F_{\beta}\uparrow$ &$\mathcal{M}\downarrow$
			&$E_{\xi}\uparrow$ &$S_{\alpha}\uparrow$ &$F_\beta^\omega \uparrow$ &$F_{\beta}\uparrow$ &$\mathcal{M}\downarrow$ \\
			\midrule

			1 & ST-SAM(Light) & 0.852 & 0.787 & 0.653 & 0.694 & 0.040 & 0.884 & 0.817 & 0.741 & 0.796 & 0.051 & 0. & 0. & 0. & 0. & 0. \\
			2 & ST-SAM(BCE-IoU) & 0.864 & 0.819 & 0.698 & 0.717 & 0.034 & 0.904 & \textcolor{red}{0.860} & \textcolor{red}{0.798} & 0.826 & \textcolor{red}{0.040} & 0. & 0. & 0. & 0. & 0. \\
			\midrule

			3 & SSCOD & 0.805 & 0.725 & 0.537 & 0.578 & 0.057 
			& 0.844 & 0.767 & 0.652 & 0.700 & 0.073 & 0. & 0. & 0. & 0. & 0. \\ 
			4 & ST-SAM & \textcolor{red}{0.884} & \textcolor{red}{0.824} & \textcolor{red}{0.713} & \textcolor{red}{0.737} & \textcolor{red}{0.032} 
			& \textcolor{red}{0.907} & 0.855 & 0.795 &\textcolor{red}{0.830} & 0.043 & \textcolor{red}{0.879} & \textcolor{red}{0.819} & \textcolor{red}{0.753} & \textcolor{red}{0.804} & \textcolor{red}{0.070} \\ 
			\bottomrule
			\hline
	\end{tabular}}
	\vspace{-\baselineskip}
	\label{tab:5}
\end{table}

\textbf{Scalability Verification.} We conducted the following experiments to verify that ST-SAM does not rely on a specific network or loss function, with the results shown in Table \ref{tab:5}, using $F_\beta ^\omega $ of COD10K as an example for comparison.

(1) Replacing the COD network with PRNet \cite{ref39}, a lightweight network designed for fully supervised COD;
(2) Replacing the loss function with BCE-IoU loss.
From the results, it can be observed that replacing the network with a lightweight one leads to a performance loss of 6\%, but still achieves an 11.6\% improvement compared to SSCOD. Meanwhile, replacing the loss function with BCE-IoU results in only a slight performance loss of 1.5\%, while maintaining a 16.1\% improvement over SSCOD. In conclusion, ST-SAM retains its low manual annotations dependency when different COD networks and loss functions are replaced, demonstrating its flexibility to adapt to various application scenarios and strong scalability.


\section{Conclusion}
In this paper, we explore the potential of SAM for self-training semi-supervised camouflaged object detection for the first time and propose ST-SAM. To move beyond the limitations of complex model design details and achieve stronger scalability and generalization, we introduce EDF to prevent error accumulation during self-training. To meet the high demand for initial models in self-training, we propose DPC to enable SAM to unleash its potential on specific tasks through hybrid prompts containing domain knowledge. Experimental results on four datasets demonstrate that ST-SAM achieves performance comparable to fully supervised and semi-supervised methods while significantly reducing the requirement for labeled samples to 1\%. We believe that ST-SAM will inject new vitality into research on SSCOD.

\section{Acknowledgments}

This work was supported by the Jilin Provincial Science and Technology Development Program (Grant No. 20220203035SF) and National Natural Science
Foundation of China under Grant (62472067).

\bibliographystyle{ACM-Reference-Format}
\bibliography{ref}

\end{document}